# Title

Data-Centric Neuromotor Interfaces for Portable Human-Machine Interaction

# Authors and Affiliations

**Jiaxuan Li[1†], Di Wu[1†], Jianhua Liu[1], Yuxin Zhao[1], Jinnuo Li[1], Xiao Zhang[1], Zhenzhi Ying[2], Changsheng Dai[3], Xiang Li[4], Liming Shu[1]**

*†These authors contributed equally to this work.*

*[1]Intelligent Equipment and Medical Device Laboratory, School of Mechanical Engineering, Dalian University of Technology, Dalian 116024, China.*

*[2]Manufacturing Laboratory, Department of Mechanical Engineering, The University of Tokyo, Tokyo 113-8656, Japan.*

*[3]Institute of Robotics and Intelligent Systems, Dalian University of Technology, Dalian 116024, China.*

*[4]Second Affiliated Hospital of Dalian Medical University, Dalian Medical University, Dalian 116024, China*

**Correspondence to:** Prof./ *Dr. Liming Shu, Intelligent Equipment and Medical Device Laboratory, Department of Mechanical Engineering, Dalian University of Technology, Dalian 116024, China.* E-mail: l.shu@dlut.edu.cn; ORCID: 0000-0002-5780-9420

# Unstructured Abstract

Dexterous human-machine interaction requires intuitive and expressive interfaces that can be efficiently deployed on constrained edge devices. Flexible material-based neuromotor interfaces hold considerable promise, as they decode human movement intention into natural control. Although emerging flexible electronic skins enable wearable high-fidelity data acquisition, practical deployment inevitably involves trade-offs between computational resources and portability. We present a data-centric paradigm where physiological features yield fundamental separability, providing sufficient discriminative cues for recognition. A wireless, high-bandwidth system developed for collecting various electrophysiological signals, when integrated with muscle-specific electrodes, forms a surface electromyography-based interface. Exploiting highly separable data, a 2,210-parameter model achieves 94.36% accuracy across 34 gestures and can be rapidly deployed on edge devices, establishing a new thousand-parameter benchmark for dexterous decoding. The underlying

data–algorithm interactions in the data-centric paradigm is further clarified, demonstrating its feasibility in real-world scenarios. This study provides a principled and validated pathway for practical deployment of reliable neuromotor interfaces.

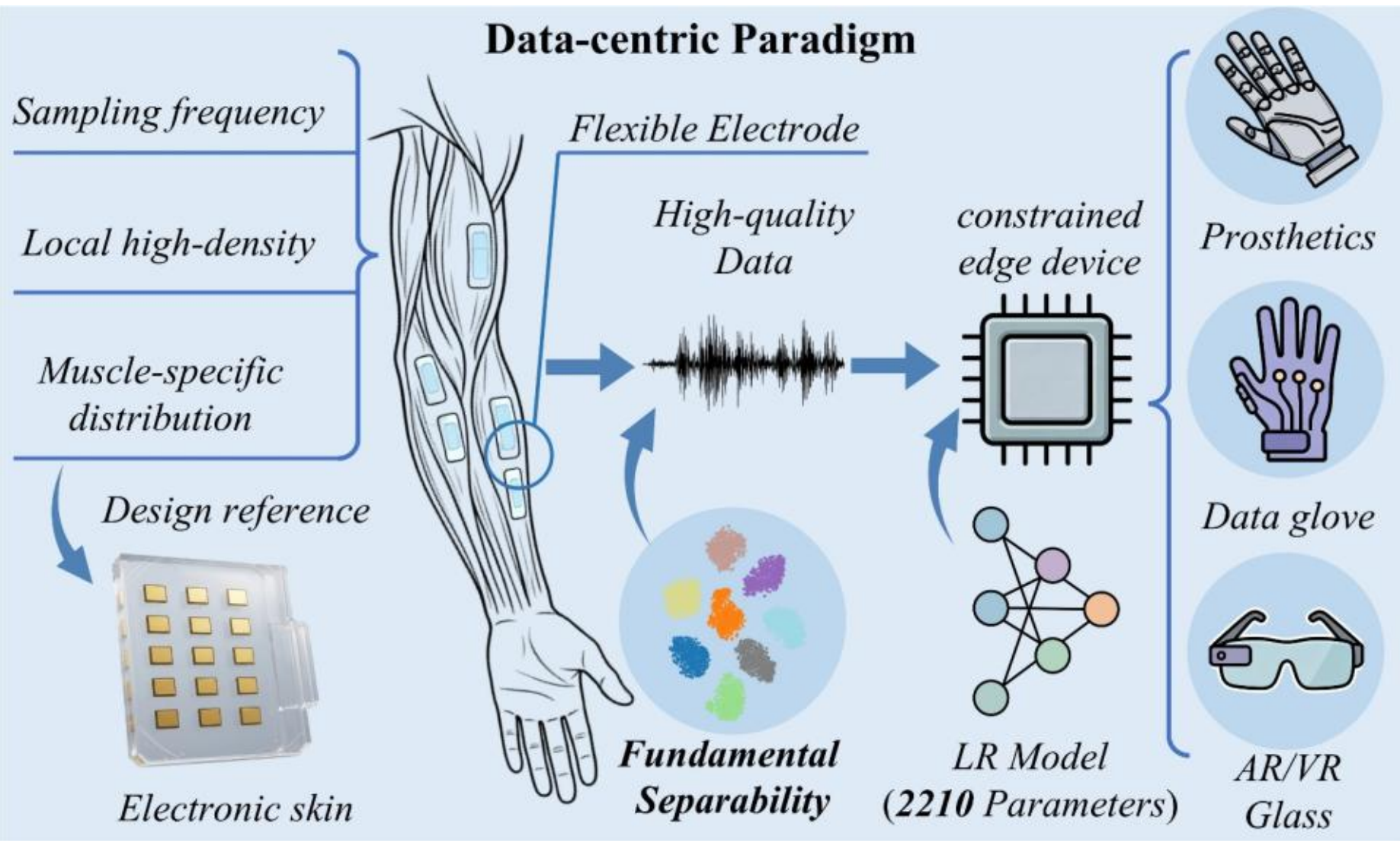

## Introduction

Neuromotor interfaces are driving a paradigm shift in human-computer and human-machine interaction (HCI/HMI) toward more portable, convenient, and compact integration[1,2]. Neuromotor interfaces decode human motion intentions from physiological signals including electroencephalography (EEG), surface electromyography (sEMG), and force myography (FMG)[3], and convert these intentions into device commands, delivering inherent capabilities of intuition, naturalness, and expressiveness. These have been demonstrating notable potential in prosthetics[4,5], rehabilitation[6–8], virtual/augmented reality (VR/AR)[9,10], and unmanned aerial vehicle control[11]. Nevertheless, most existing designs face challenges in resource-constrained deployment scenarios.

A principal obstacle is the prevailing model-centric philosophy, which prioritises decoding accuracy through increasingly sophisticated mathematical models[12,13], machine learning [14], and deep learning models[15–19].

Consequently, there is a tendency to adopt intricate algorithmic architectures for diverse and complex intent decoding tasks. However, excessively intricate algorithmic architectures are adopted to handle diverse and complex intent decoding tasks. The massive parameter scale exceeds hardware capacity, while sophisticated computational strategies are incompatible with hardware acceleration. As a result, this pursuit of performance yields architectures with prohibitive computational overhead[20], fundamentally conflicting with edge device constraints and ultimately hindering practical deployment.

In contrast to the model-centric paradigm, a data-centric philosophy pursues efficient model construction through enhanced data quality, consistency, and coverage[21]. This philosophy has achieved substantial progress in partial HMI issues. For instance, by expanding training data to 4900 participants, relevant research realises stable, high-accuracy cross-user gesture recognition meeting commercial application standards[22]. Despite the notable advancement, the application of this paradigm to neuromotor interfaces presents unique challenges, mainly the high cost of large-scale data acquisition. An alternative data-centric pathway prioritises improving data quality to endow original data with distinct intrinsic spatiotemporal structures and separability, enabling simultaneous optimisation of performance and computational efficiency without trade-offs. Accordingly, owing to the excellent deformability of materials that mitigates signal quality deterioration induced by muscle deformation, research on flexible electronic skins and wearable devices is developing rapidly[23].

However, systematic research on physiological characteristics based on electrode array electrical parameter configurations remains insufficient, leaving a lack of practical parameter references for various application scenarios. Besides, developing easily deployable, high-precision, and portable designs for HMI-oriented neuromotor interfaces, as well as exploring the correlation between data quality and algorithmic complexity for stable and efficient neuromotor decoding, are still pending research objectives.

In this study, we further explore the data-centric paradigm in the design of HMI-oriented neuromotor interfaces, using forearm sEMG as the physiological input signal for gesture recognition. Crucially, our primary novelty in neurobiological interfaces lies neither in proposing another high-density sensor nor in designing lightweight algorithms, but in establishing a core principle that physical spatio-temporal fidelity inherently simplifies feature geometry, thereby allowing ultra-compact linear models to achieve accurate intent recognition without heavy neural backbones. We developed a neuromotor interface to validate the feasibility of a data-centric approach systematically. Leveraging the collected high-quality data, an ultra-lightweight model with merely 2,210 parameters (17.68 kB) realises stable and high-accuracy gesture recognition on resource-constrained edge devices. It reaches an online accuracy of 99.14% for 34 gestures and 98.99% overall accuracy in the 20-gesture robotic hand control task, which transcends the conventional optimisation strategy of model-centric methods dependent on structural redesign and parameter expansion. More importantly, we systematically investigated the effects of diverse electrical parameters on signal quality and physiological characteristic fidelity captured via flexible electrode arrays. The results reveal that it is this prominent fundamental separability of high-quality signals that enables lightweight algorithms to achieve superior recognition performance [Figure 1A]. Here, fundamental separability describes the inherent discriminability among feature distributions of different motion categories directly within raw high-fidelity physiological signal space. We deduce that such characteristics originate from well-defined spatiotemporal patterns embedded in high-fidelity physiological recordings. The neuromotor interface incorporates a novel 64-channel distributed electrode array (D64) and an innovative wireless multi-channel bioelectrical signal acquisition unit, supporting acquisition of diverse physiological signals [Figure 1B]. These results demonstrate the feasibility of applying the data-centric design philosophy to HMI-oriented neuromuscular interfaces, overcoming the limitations of model-centric approaches and providing a simple, adaptable, and scalable strategy, thereby establishing a roadmap toward highly intuitive and broadly applicable, edge-deployable HMI.

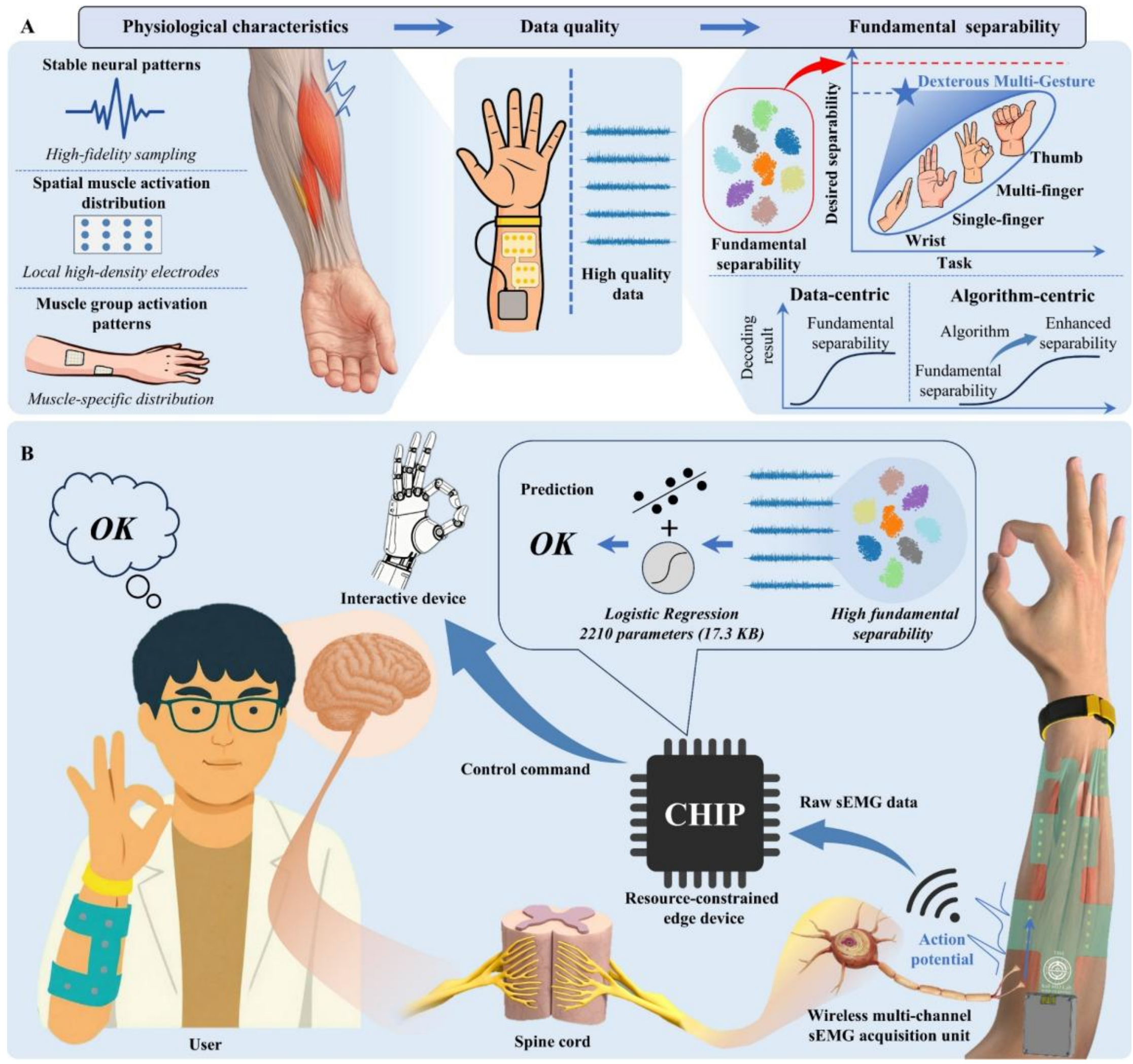

**Figure 1.** The Developed Interface Inspired by the Data-Centric Paradigm. (A) Physiological signals with sophisticated spatiotemporal characteristics possess remarkable fundamental separability, driving low-complexity algorithms to achieve high-precision recognition. (B) The developed interface workflow. Sufficient fundamental separability supports reliable gesture recognition on resource-constrained edge devices for real-time interaction.

# Materials and Methods

## sEMG Acquisition Unit

The developed wireless acquisition unit [Figure 2A] uses a common-ground differential sampling architecture, with a wristband electrode as a reference (Spes Medica S.p.A., Italy). This unit can capture diverse human physiological signals, including sEMG, electrocardiography (ECG), EEG, and electrooculography (EOG). Capable of 64-channel wireless signal acquisition at 2000 Hz, the device adopts a pluggable electrode connection design. All channels exhibit a baseline noise below 1.5 μVrms. The device measures 52 × 42 × 13.4 mm and weighs 32.4 g. It can sustain up to four hours of continuous runtime, powered by a built-in battery and power management unit (PMU). Three functional modules embedded in the front-end chip, namely differential amplification, bandpass filtering, and analogue-to-digital conversion, support stable physiological signal acquisition.

**Data-centric HMI-oriented neuromotor interface**

An HMI-oriented neuromotor interface was built, based on the data-centric paradigm [Figure 1B]. Upon issuing motor commands, the brain transmits electrical impulses along nerves to muscles and triggers muscle fibre contraction. Partial bioelectrical activity propagates to the skin surface, generating sEMG signals. These signals are acquired via a 64-channel electrode array and wirelessly forwarded to edge hardware. The device extracts signal features and recognises target gestures with pre-trained algorithms. Recognition results are converted into control signals and delivered to actuators like robotic hands.

The integrated interface consists of the wireless acquisition module and D64 electrodes. The D64 electrode array is tailored to the anatomical distribution of forearm muscles [Figure 2B], while a 64-channel ring electrode array (R64) adopts a conventional circular arrangement [Figure 2C]. Both electrodes match the acquisition unit, and the R64 array serves as a reference for comparative analysis against the D64 array. Compared with the R64 electrode, the channels of the D64 electrode are more widely distributed, leading to greater signal variation between channels [Figure 2D, Figure 2E].

**Gesture Group and Electrode Design**

A total of 34 gestures were included in this study, comprising one relax gesture and four groups of movement gestures: single-finger, multi-finger, wrist, and thumb [Figure S1]. The adopted gesture set covers tasks of varying recognition difficulty that correspond to daily functional movements, including easily distinguishable single-finger and wrist motions with a single degree of freedom, as well as challenging multi-finger and thumb fine motor gestures that require sophisticated motion control and involve higher degrees of freedom [Figure 1A][24]. These gestures engage nearly all major muscles of the forearm [Figure S2A].

Located deep in the palmar forearm, the flexor digitorum superficialis and flexor digitorum profundus play a crucial role in gesture recognition, which primarily control flexion of all fingers except the thumb. Located around the elbow, radial and ulnar forearm muscles such as flexor carpi radialis, palmaris longus, and flexor carpi ulnaris feature large cross-sectional areas, yielding intense sEMG signals. In contrast, muscles like the extensor pollicis longus, extensor pollicis brevis, and extensor indicis possess narrow distal bellies and exhibit considerable spatial overlap with limited surface exposure around the wrist. Therefore, electrode channel placement must carefully consider both muscle distribution and involvement in hand movements, balancing deep signal extraction with wide-area coverage to optimise accuracy and stability in dexterous multi-gesture recognition. Based on the above factors, the D64 electrode array was designed in accordance with actual muscle distribution [Figure 2B]. The channels are divided into five regions. Region A covers the flexor carpi radialis, palmaris longus, flexor carpi ulnaris, and the deep flexor digitorum superficialis and flexor digitorum profundus. Given the depth and functional significance of the flexor digitorum muscles, this region is allocated 20 channels to enhance signal capture density. Region B covers the extensor carpi radialis brevis, extensor carpi ulnaris, and extensor digitorum, representing most of the superficial dorsal forearm muscles. Since the extensor digitorum plays a prominent role in extension movements, 24 channels are assigned to this region to improve recognition performance. Region C collects signals from the extensor digiti minimi, abductor pollicis longus, and the majority of extensor digitorum muscles, with 12 channels allocated. Regions D and E jointly cover the extensor pollicis longus, extensor pollicis brevis, and extensor digitorum indicis, which have limited surface exposure near the proximal wrist. For all regions, electrode spacing is 10 mm along the forearm axis and 20 mm around its circumference.

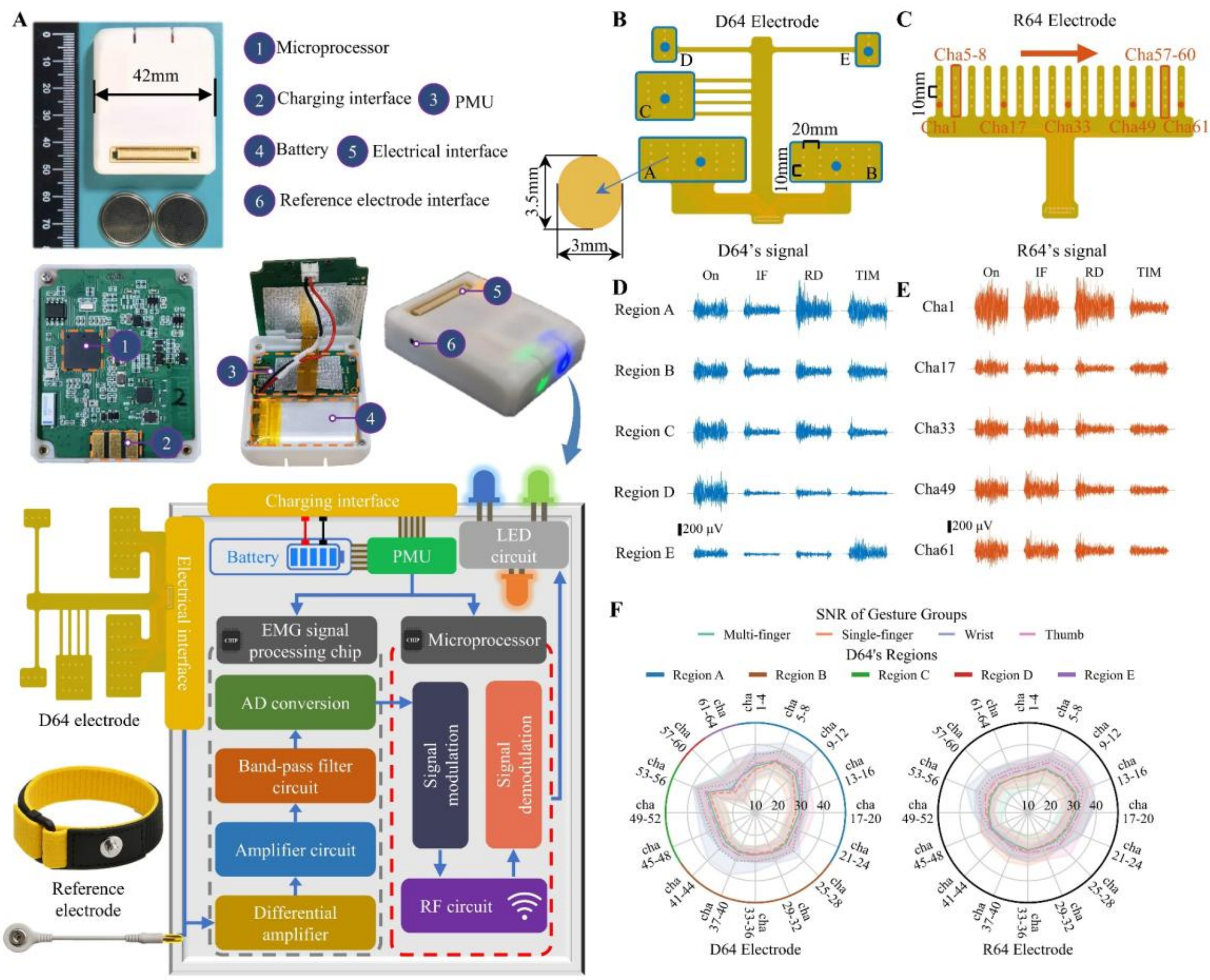

**Figure 2.** Wearable Bioelectrical Signal Acquisition Unit and sEMG Electrode Arrays. (A) Schematic diagram and photograph of the sEMG acquisition unit, showing both the internal circuitry and the external appearance. (B) Electrode layout of the D64 array. (C) Electrode layout of the R64 array. (D) Example sEMG signals recorded from selected electrodes (blue points) in (B). (E) Example sEMG signals recorded from selected electrodes (red points) in (C). (F) Illustration of the SNR of the R64 and D64 arrays. Gestures are divided into four groups for visualisation, and the D64 array additionally includes colour-coded regions corresponding to different muscle groups.

To provide a comparison with the traditional ring-shaped layout [24–27], the R64 electrode array [Figure 2C] was designed. The array adopts a 4×16 layout, arranged in four rows along the forearm axis and sixteen columns along the circumferential direction [Figure S2B, Figure S2C]. Electrode spacing is 10 mm along the forearm axis. The comb-like grouped electrode design adapts to the tapered forearm contour, enabling snug attachment and steady signal capture. The array is available in four circumference sizes [220 mm, 225 mm, 230 mm, 235mm] to accommodate users with different forearm dimensions.

An ultrathin polyimide film serves as the electrode substrate. The custom-designed D64 distributed electrode array and R64 ring electrode array were both manufactured using commercial flexible printed circuit fabrication services (JLCPCB, Shenzhen, China) with a thin gold surface finishing layer deposited on each electrode pad. The electrodes were attached to the skin using high-density foam double-sided adhesive pads (Spes Medica S.p.A., Italy) with pre-punched micro-apertures corresponding to the electrode pads, and conductive paste (AC Cream, Spes Medica S.p.A., Italy) was applied to the electrode sites.

Given their identical materials and manufacturing processes, the D64 electrode served as a representative benchmark for characterisation. Although the polyimide substrate is non-stretchable, its ultrathin profile provides high flexibility and conformability, allowing it to seamlessly follow skin contours. The trace impedance from the electrode site to the interface connector was measured using a precision impedance analyser (6500B, Wayne Kerr Electronics, UK). The resulting impedance remains extremely low, around 1 Ω, and completely flat from 20 Hz to 10 kHz, indicating high structural conductivity and minimal signal loss [Figure S3A]. Furthermore, in vivo skin–electrode impedance was evaluated across four repeated attach–detach trials using two sites spaced 5 cm apart. The resulting curves show classic capacitive decay and near-perfect overlap across all trials, confirming excellent repeatability and reusability [Figure S3B]. Compared to recent sustainable interfaces such as the silk fibroin ion-aerogel microstructured dry electrode [28], the D64 electrode achieves highly comparable skin–electrode contact impedance while offering robust conformal contact and reliable signal transmission. To accommodate dynamic skin deformations from movement, the flexible polyimide substrate works with the medical adhesive to maintain tight physical contact, effectively preventing motion-induced delamination. Finally, all components, including the polyimide substrate, gold layers, and conductive paste, are biocompatible and non-irritating, ensuring safe and long-term continuous wear for extended neural decoding.

We calculated the SNR for each gesture with both electrode types and summarised the results according to four groups: single-finger, multi-finger, wrist, and thumb movements [Figure 2F, Table S1]. The SNR was calculated using the following formula:

$$\mathrm{SNR\ (dB)} = 20 \cdot \log_{10}\left(\frac{\mathrm{RMS_{signal}}}{\mathrm{RMS_{noise}}}\right). \qquad (1)$$

Here, $RMS_{signal}$ represents the Root Mean Square (RMS) value, obtained from the sEMG samples in the gesture segment for that channel, and $RMS_{noise}$ is the RMS value measured during rest.

The D64 array presents notable inter-channel discrepancies, predominantly reflecting activation of extensive muscle groups and yielding prominent spatial differentiation. In contrast, the R64 array delivers relatively consistent signal amplitude among channels and gesture patterns, with weaker inter-channel variation and a narrower dynamic range of recorded muscle activity[29]. In summary, the D64 configuration captures spatially localised muscle activation, while the R64 layout generates evenly distributed activation signals across all channels.

## Data Preprocessing and Feature Extraction

All data processing procedures, including filtering, feature extraction, and classification, are implemented on edge devices in sequence. Preprocessing is required as raw data contains task-irrelevant noise. Since the main energy of sEMG signals is concentrated between 20 and 500 Hz, the raw signals are first processed with a 20–500 Hz bandpass filter. A notch filter with a quality factor of 50 is then applied to attenuate 50 Hz power-line interference and its harmonics.

Root Mean Square (RMS) features were extracted from all channels, calculated with a 200 ms sliding window and 100 ms step size. For a windowed sample containing T data points, the RMS feature for the $j$-th channel is defined as follows:

$$\mathrm{RMS}_j = \sqrt{\frac{1}{\mathrm{T}}\sum_{i=1}^{T} x_{i,j}^2}. \tag{2}$$

The obtained features were normalised via Z-score and then mapped to predefined gesture categories using the Logistic Regression (LR) classifier. Based on the recognition results, the corresponding control commands are transmitted to the interaction devices.

## Data Collection

This study recruited 15 healthy volunteers, including 11 males and 4 females, all right-handed. All participants had no history of muscular or neurological disorders and provided written informed consent. The experimental protocol was approved by the Ethics Committee of Dalian University of Technology (approval number: DUTSME230406-01) and was conducted in accordance with the Declaration of Helsinki.

sEMG signals were collected from all participants. Before the formal test, all participants received 15 minutes of training to get acquainted with gestures. Seven participants were tested with the D64 array first and then the R64 array, whereas the other eight adopted the opposite test order. Each gesture was held for 4 seconds, with a 5-second rest interval afterwards. Every participant finished 11 experimental sessions, and 34 gestures were carried out in fixed order within each session. Data were grouped by session. To strictly evaluate model performance and prevent data leakage, 1 independent session was held out as a validation set for early stopping and hyperparameter monitoring. The remaining 10 sessions were employed in a 10-fold session-based leave-one-out cross-validation scheme, where each fold utilized 9 sessions for training and 1 session for testing. The final recognition performance of each subject was calculated by averaging results from all ten test folds.

### HMI Offline Performance Evaluation

The physiological features of the sEMG signal include stable neural patterns, muscle group activation patterns, and localised muscle activity. These features can be controlled through temporal sampling rate (TSR) [Figure 3A] [30], spatial sampling rate (SSR) [Figure 3B] [31] across recording sites, and channel layout (CL) [Figure 3C] [32–34]. To investigate their effects on data quality and decoding performance, we established baseline variables and systematically manipulated TSR, SSR, and CL [Figure 3D]. For the baseline, we selected eight circularly arranged channels at 200 Hz, consistent with the signal quality settings used in commercially available sEMG-based HMI devices such as the Myo armband[35]. Data preprocessing was implemented as illustrated in [Figure 3D]. Raw signals sampled at 2000 Hz were downsampled to four temporal sampling rates: 1000 Hz, 500 Hz, 400 Hz, and 200 Hz. Together with the original 2000 Hz signal, a total of five TSR levels were evaluated. Channel selection was conducted for both D64 and R64 electrode arrays. Let n denote the channel count along the longitudinal direction of the arm, ranging from 1 to 4, and let m represent the circumferential channel number. Eight spatial sampling configurations were generated. Together with two channel layout schemes, 80 distinct data property combinations were formed and evaluated using seven machine learning classifiers. Each dataset was defined by unique parameters covering sampling rates, channel layout, classification algorithm, and experimental subject.

To investigate whether high-quality data can drive low-complexity algorithms to achieve high-precision dexterous gesture recognition, seven classical machine learning classifiers with low computational overhead were employed: Decision Tree (DT), k-Nearest Neighbours (KNN), Random Forest (RF), Linear Discriminant Analysis (LDA), Support Vector Machine (SVM), Multi-Layer Perceptron (MLP) and LR. DT recursively partitions the feature space according to threshold conditions until each sample is mapped to a class label. KNN assigns the class of a sample based on the majority label among its K closest neighbours in the feature space. RF enhances the robustness of DT by constructing multiple trees on randomly selected feature subsets and aggregating their outputs through ensemble voting. LDA projects samples into a linear subspace that maximises inter-class separability while minimising intra-class variability. SVM constructs an optimal hyperplane to maximise the margin between classes and employs kernel transformations for nonlinearly separable data. MLP leverages multiple layers of interconnected neurons with nonlinear activation functions to learn complex mappings between features and class labels. LR estimates class probabilities via a logistic function applied to a linear combination of features. It requires only a matrix multiplication and a bias addition, followed by selecting the class with the highest output, making it the simplest classifier with the fewest parameters. For non-iterative classifiers (DT, KNN, RF, LDA, and SVM), hyperparameter tuning was conducted via grid search with default convergence solvers. For iterative models (MLP and LR), training was configured for a maximum of 5,000 iterations using an early stopping criterion based on the held-out validation session. Model performance on this validation session was assessed at 50-iteration intervals. Training was effectively terminated when the validation loss plateaued, which typically occurred around 1,200 iterations, and the optimal model checkpoint was saved for final evaluation.

For each dataset, the overall classification accuracy for multi-class recognition was calculated using:

$$\text{Accuracy} = \frac{\sum_{i=1}^{K} N_{i,i}}{\sum_{i=1}^{K}\sum_{j=1}^{K} N_{i,j}} \tag{3}$$

where $K$ denotes the total number of gesture categories, $N_{i,j}$ represents the number of samples from class $i$ predicted as class $j$, and $N_{i,i}$ denotes the number of correctly classified samples for class $i$. The Wilcoxon signed-rank test was adopted for all statistical significance analyses. For clarification, every single metric and quantitative calculation throughout the entire manuscript was strictly derived using all 34 classes.

**HMI Online Performance Evaluation**

To evaluate the feasibility of our HMI-oriented neuromotor interface, we conducted two online experiments: an online gesture recognition test covering all 34 gestures, and a robotic hand control test involving 20 gestures using a dexterous robotic hand (Inspire-Robots, Beijing, China) due to its degrees-of-freedom limitations. For edge deployment, we adopted the ESP32-S3 (Espressif Systems, Shanghai, China), a high-performance AIoT system-on-chip integrating a 2.4

GHz Wi-Fi and Bluetooth 5 (LE) dual-core MCU (Xtensa LX7, up to 240 MHz), 512 KiB SRAM, and rich peripheral interfaces, which is widely used in embedded machine learning and real-time HMI applications. Model training was performed on a workstation equipped with a 12th Gen Intel(R) Core(TM) i5-12600KF CPU. The LR model was trained on a central computer using Python 3.10.13 and scikit-learn 1.6.1, before being deployed onto ESP32-S3. In offline validation, the training and validation trajectories [Figure S4] closely mirror each other throughout convergence, showing no signs of overfitting under this setup.

### Analysis of Variance

To formally assess the impact of each data property and subject on classification accuracy under different algorithms, we employed a multiple linear regression model with all factors treated as categorical variables. The model included main effects, two-way interactions, and three-way interactions:

$$Y = \beta_0 + \sum_i \beta_i X_i + \sum_{i<j} \beta_{ij} X_i X_j + \sum_{i<j<k} \beta_{ijk} X_i X_j X_k + \epsilon, \quad (4)$$

where $Y$ denotes classification accuracy, $X_i$ represents the three data properties and subject, and $\in$ is the residual error.

To quantify the effect size of each factor and their interactions, $\omega^2$ was computed. $\omega^2$estimates the proportion of variance explained by a component while accounting for residual error, providing a less biased measure than partial $\eta^2$, particularly when sample sizes are limited. It is calculated as

$$\omega^2 = \frac{SS_{\text{Effect}} - df_{\text{Effect}} \cdot MS_{\text{Error}}}{SS_{\text{Total}} + MS_{\text{Error}}}, \quad (5)$$

where $SS_{Effect}$ is the sum of squares for the component or interaction, $df_{\text{Effect}}$ is its degrees of freedom, $MS_{Error}$ is the residual mean square, and $SS_{Total}$ is the total sum of squares. Higher values of $\omega^2$ indicate a stronger impact of the factor on classification accuracy.

## Results and Discussion

### High-Quality Data Empowers Low-Complexity Algorithms

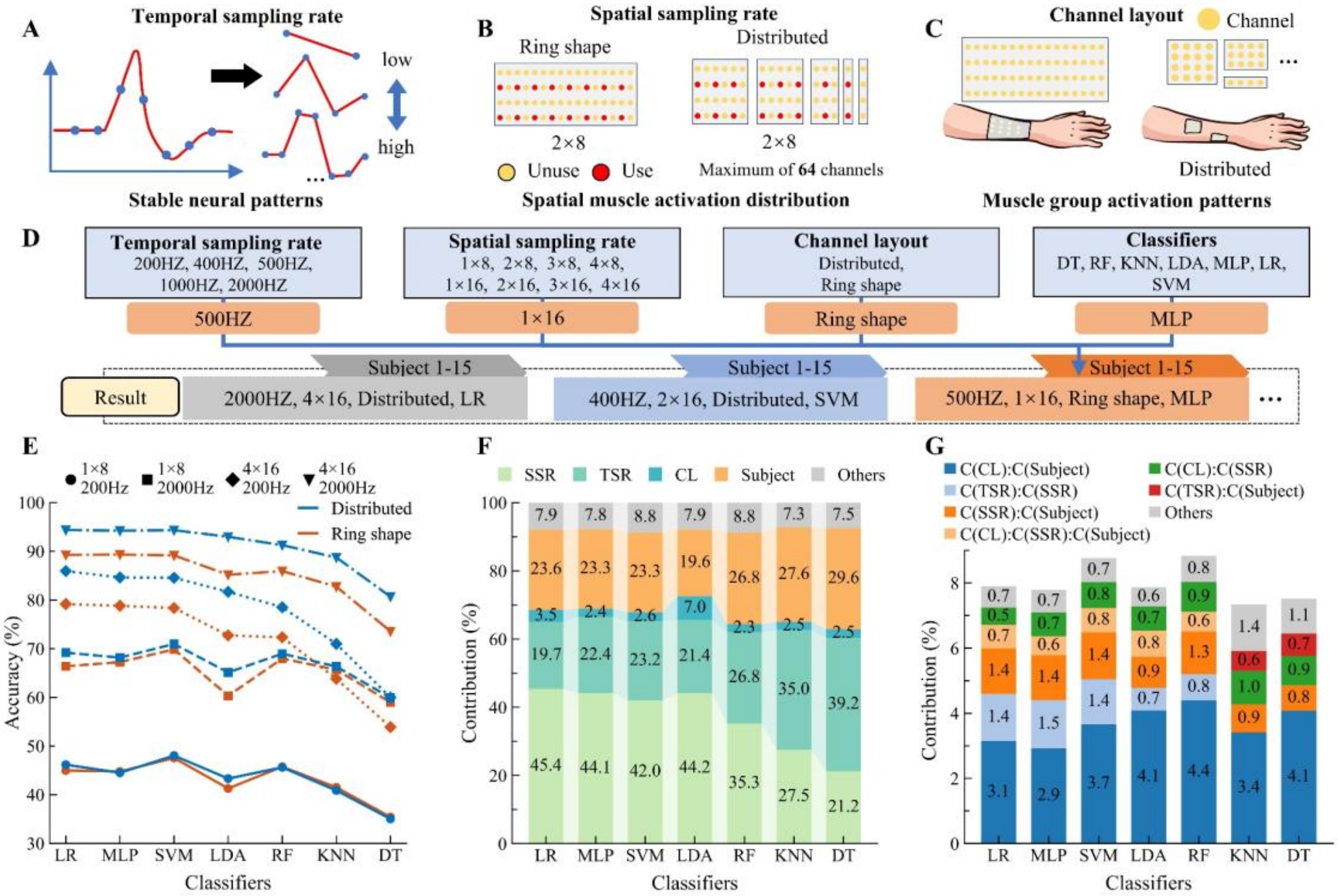

**Figure 3.** The variation of the recognition accuracy of low-complexity algorithms with different signal quality. (A–C) Schematics of experimental settings for temporal sampling rate, channel layout, and spatial sampling rate. (D) Evaluation of recognition accuracy under different combinations of data properties. (E) Recognition accuracy under various data properties. Higher data quality enables high accuracy even with low-complexity algorithms. ANOVA results for main effects (F) and higher-order interactions (G), identifying TSR and SSR as dominant factors.

Classification accuracy under diverse data configurations illustrates the correlation among signal quality, decoding performance, and classifier characteristics [Figure 3E]. First, rising TSR [Figure S5A] and SSR [Figure S5B] both boost decoding capability, while the performance gain of CL depends on SSR conditions [Figure S5C]. Second, variations in data parameters alter decoding accuracy and the comparative ranking of algorithms. TSR improvement brings comparable performance gains to all classifiers [Figure S5D]. In contrast, SSR exerts disparate effects on different algorithms [Figure S5E]. Apart from LDA, CL delivers roughly equal performance enhancement across the remaining models [Figure 3E]. Thus, SSR substantially influences the relative ranking of the evaluated classification algorithms under these sEMG configurations. This suggests that algorithmic superiorities can be sensitive to specific data setups, underscoring the necessity of multi-scenario validation [Figure S5F]. Apart from this case, all algorithms achieved substantial accuracy gains. Results show that high-quality data paired with simple algorithms yields excellent decoding performance, providing a practical and stable solution for deploying lightweight algorithms on edge devices.

Analysis of variance was adopted to assess how signal quality factors affect experimental outcomes [Figure 3F, Figure 3G]. TSR and SSR exert the most prominent influence on classification accuracy. This indicates that steady neural characteristics and subtle local muscle activations play a predominant role in motion intention decoding within this framework. Meanwhile, individual variation in CL accounted for approximately 4% of the variance, suggesting that fine-tuning electrode configurations based on individual anatomical differences may offer additional decoding refinements.

**Data Separability Determines Decoding Capability**

The selected algorithms are considerably lightweight compared to state-of-the-art deep learning models. For instance, the LR model merely contains 2,210 parameters, yet it surprisingly attained optimal performance in offline tests. The LR model is an extremely shallow structure, lacking powerful capability for deep feature extraction and fusion. Classification cues therefore stem almost entirely from the raw data, indicating that the primary determinant of classification performance is the inherent fundamental separability of the data. High-quality physiological signals, once feature-extracted, can yield highly separable data[36].

Based on this speculation, the intrinsic data separability is further explored and visualised. Three data properties were selected with two levels each, including TSR of 200 Hz and 2000 Hz, SSR of 1×8 and 4×16, and CL of D64 and R64, which were combined into eight cross-configurations. t-Distributed Stochastic Neighbour Embedding (t-SNE) is adopted for visual presentation. To systematically analyse signal distribution, the 34 gesture categories were categorised into five logical groups: four active gesture groups (single-finger, multi-finger, wrist, and thumb gestures) and one relax gesture. However, due to its near-zero sEMG amplitude and substantially different data distribution from active muscle contractions, visualising the Re gesture as an isolated category provides limited insight, and it was thus excluded from explicit labelling in the t-SNE projection. Since 33 gesture categories were considered, substantial overlap occurred during projecting them into the same 2D space. To address this, the gestures were grouped into four categories: single-finger, multi-finger, wrist, and thumb gestures. As illustrated in [Figure 4A], distinct boundaries can be observed between single-finger and wrist gestures under the 2000 Hz, 1×8 channel arrangement. In comparison, unambiguous separation of multi-finger and thumb gestures is only achieved under the distributed 2000 Hz, 4×16 configuration. Overall, optimising TSR, SSR, and CL gradually forms obvious data clustering based on spatiotemporal discrepancies. Such inherent differences can be effectively captured by shallow machine learning models without complex feature fusion modules, laying a theoretical foundation for lightweight deployment.

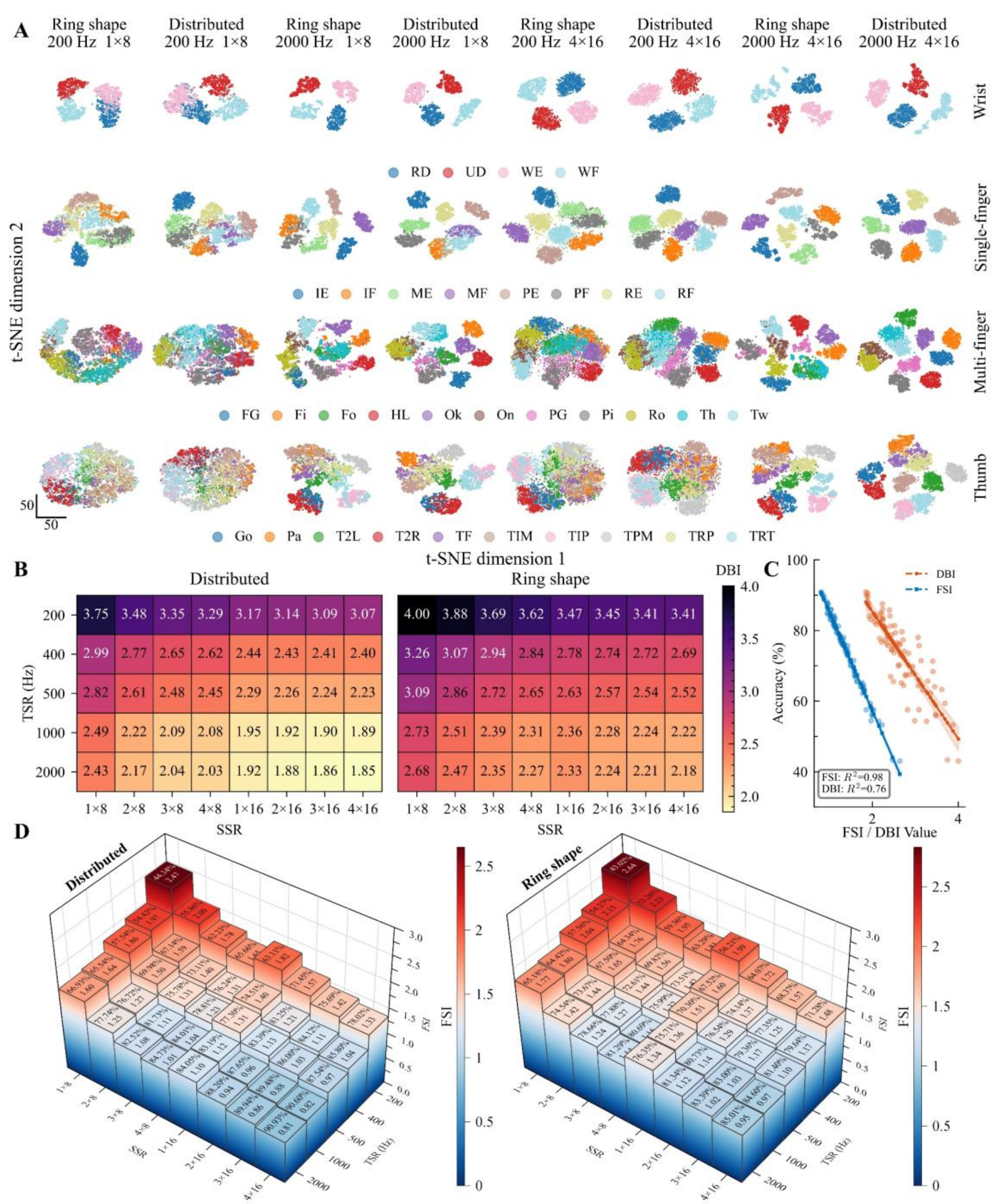

**Figure 4.** Enhanced data quality directly gives rise to clear separability in the feature space. The t-SNE visualization (A) reveals that progressively improved data quality leads to increasingly clear class boundaries among different gesture groups. The monotonic decrease in DBI (B) further supports this trend. However, its correlation with recognition accuracy is relatively weak. In contrast, the FSI (C) exhibits a stronger correlation with accuracy, enabling more effective quantification of inherent separability (D).

We further speculate that under the specified high-fidelity configuration, the three key factors of signal acquisition play different and complementary roles in the stability and diversity of physiological signal representations. Raised TSR enhances the fidelity of acquired physiological signals, facilitating steady capture of inherent physiological fluctuations and strengthening intra-class clustering. Optimised CL regularises activation patterns, diminishes session-to-session discrepancies of confusing gestures, and improves intra-class similarity. By comparison, higher SSR yields finer spatial signal details. Nevertheless, insufficient signal fidelity brings erratic fluctuations, which enlarges inter-class divergence while raising intra-class dispersion. The results reveal that TSR and CL mainly reinforce the stability and uniformity of physiological features, while SSR promotes feature diversity and discriminability. The three factors function complementarily, shaping the representational structure of sEMG signals.

These findings offer clear guidance for deploying lightweight models in resource-limited scenarios. Priority should be given to improving the spatiotemporal fidelity of raw data, instead of merely pursuing higher space utilisation of models on edge devices. The strategic enhancement of signal fidelity and spatial detail is essential to elevate the fundamental separability of the sEMG signal. Ultimately, optimising data characteristics improves feature separability, which sets the upper limit of decoding performance for lightweight algorithms and validates the practicability of the data-driven research paradigm.

Furthermore, rigorous quantification of this separability capability is essential. Traditionally, the Davies–Bouldin Index (DBI) has been used to partially evaluate the inter-class discriminability of data, where smaller values indicate better separability [37]. For each gesture category, the intra-class dispersion $S_i$ is defined as

$$S_i = \frac{1}{G_i} \sum_{k=1}^{G_i} \| x_i^k - c_i \|, \quad (6)$$

where $G_i$ is the number of samples in gesture category $i$, $x_i^k$ is the $k$-th sample, and $c_i$ is the centroid of category $i$. The DBI is then computed as:

$$\mathrm{DBI} = \frac{1}{K} \sum_{i=1}^{K} \max_{j \neq i} \frac{S_i + S_j}{\| c_i - c_j \|}, \quad (7)$$

where $K$ is the number of gesture categories. As shown in [Figure 4B], the DBI values of sEMG features under different data property configurations, it is evident that increasing TSR and SSR, as well as optimising CL, consistently lowers DBI values contribute to progressively improved separability. These results indicate that more faithful physiological signals, richer fine-grained information, and broader-scale physiological patterns enhance signal quality, which in turn improves the intrinsic separability of the signals.

To minimize bias from any single classifier, we measured data separability using the mean accuracy of seven standard shallow models across various data configurations. Raw DBI roughly mirrored classifier accuracy [Figure 4C], but its linear fit was poor ($R^2 = 0.76$). For instance, setup configurations with low TSR and high SSR achieved higher accuracy yet yielded larger DBI values than high-TSR and low-SSR setups. Adding more channels captures richer spatial muscle patterns and expands the feature space, which helps separate data in higher dimensions. Because of this, raw DBI struggles to compare separability across feature spaces with different dimensions.

To adjust for this dimension bias, we empirically introduced a channel-scaling factor based on the fifth root of the channel count to define the Fundamental Separability Index (FSI):

$$\mathrm{FSI} = \frac{\mathrm{DBI}}{\sqrt[5]{C}}. \quad (8)$$

FSI aligned far better with classification accuracy than raw DBI (Linear regression: $R^2 = 0.985, p < 0.001$ ; Spearman correlation: $\rho = -0.992, p < 0.001$) offering a practical, classifier-agnostic proxy for estimating relative separability in standard machine learning models [Figure 4C, Figure S6]. Here, the power $n = 5$ was empirically chosen through a sensitivity analysis across various values ( $n \in [2,10]$ ). While the root-scaling strategy remained consistently stable across all tested powers (Spearman correlation: $\rho \leq -0.934, p < 0.001$; Linear regression: $R^2 \geq 0.881, p < 0.001$), $n = 5$ yielded the optimal fit [Figure S6]. As data configurations improved, FSI dropped steadily while accuracy climbed [Figure 4D]. Overall, these results show that FSI serves as a handy, empirical tool to gauge relative data separability across varying channel counts in our sEMG gesture tasks. All offline metrics were averaged across all 34 gestures to keep the analysis consistent.

**Deployment Validation**

To verify the feasibility and low computational consumption of the developed interface, an online gesture recognition experiment was carried out [Figure 5A] using the ESP32-S3 as the core computing unit. The evaluation was conducted on a representative subject, including 1 trial for the full 34-gesture dataset and 5 trials for the 20-gesture robotic hand control task. Rather than merely pursuing high accuracy, the primary objective was to validate the effectiveness of our data-centric methodology on resource-constrained edge devices. To mitigate errors during rapid gesture transitions, a transition-smoothing post-processing strategy was implemented [Figure 5B], where for transitions between actions or from rest to an action, the output gesture updates only when two consecutive sliding windows yield identical predictions; otherwise, the previous prediction is retained. The sliding window step size during the online experiment was also set to 200 ms. Exceptionally, if the latest prediction is the rest state, it is executed immediately without delay. All quantitative metrics were statistically calculated at the sliding-window frame level after applying this post-processing strategy.

Considering the uneven sample distribution of gestures in online tests, macro-averaged accuracy (MAA) was adopted for balanced category evaluation. Different from overall accuracy that is biased toward categories with more samples, MAA equally weights all gesture classes by averaging the independent accuracy of each category. Its formula is expressed as:

$$MAA = \frac{1}{K}\sum_{i=1}^{K}\frac{N_{i,i}}{\sum_{j=1}^{K}N_{i,j}} \tag{9}$$

where $K$ is the total number of gesture categories, $N_{i,j}$ represents the number of samples with true label $i$ predicted as $j$, and $N_{i,i}$ indicates the number of correctly classified samples of class $i$.

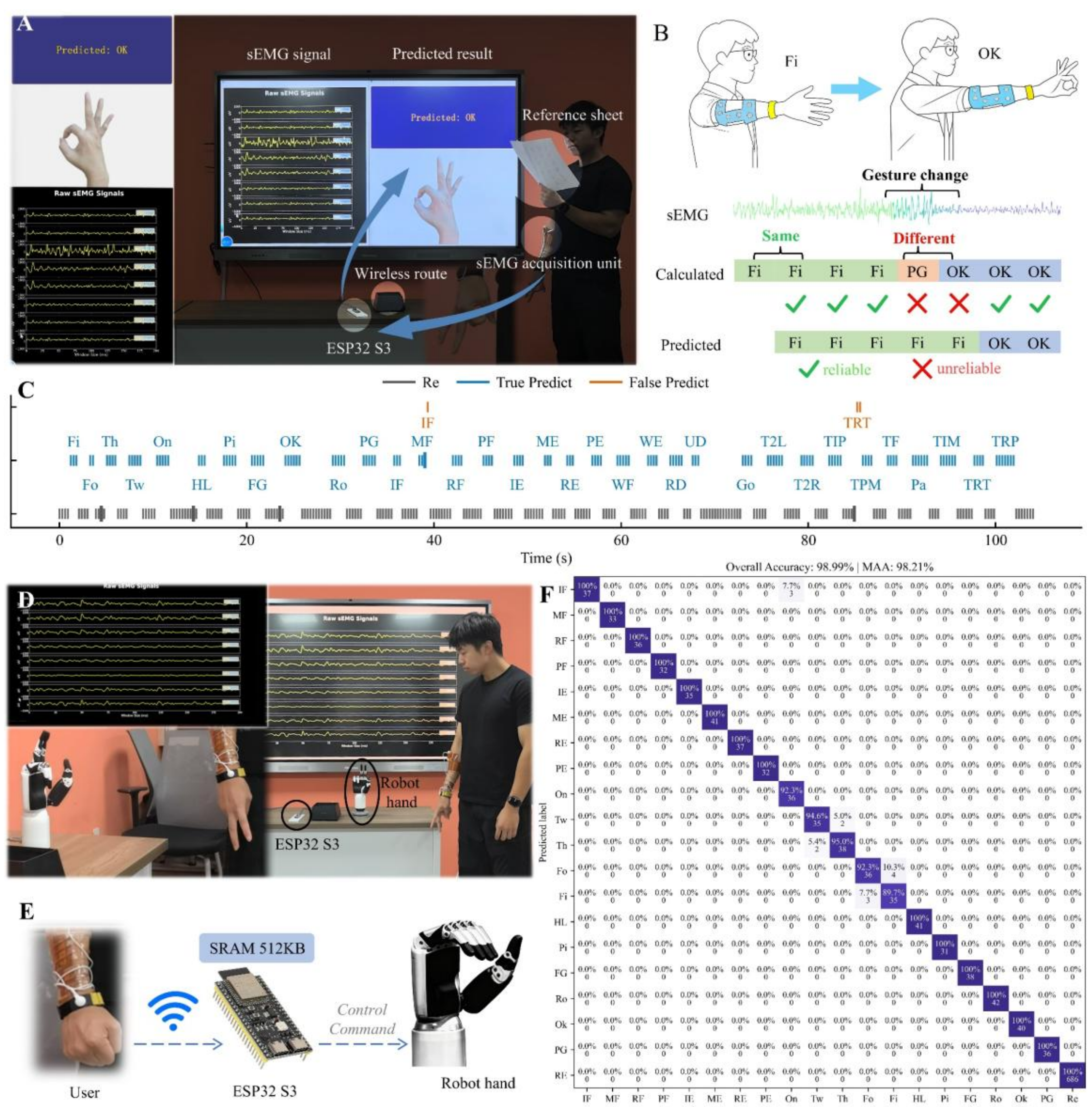

Figure 5. Data-centric paradigm enables effective edge deployment for on-site gesture recognition and robotic hand control. An online recognition experiment with 34 gestures is first conducted. The experimental setup and module communication are shown in (A), where a post-processing strategy updates predictions only when consecutive windows are consistent (B), ensuring stable online recognition results (C). Thick non-red lines denote windows successfully corrected by post-processing. Subsequently, 20 gestures are selected according to the degrees of freedom of the robotic hand for real-time control (D), supported by the communication architecture among control modules (E). The overall robotic hand control accuracy reaches 98.99%, with an MAA of 98.21% (F).

Under this scheme, the online 34-gesture recognition test [Figure 5C, Movie S1] achieved an overall accuracy of 99.14% with a corresponding MAA of 98.28%. Specifically, the post-processing strategy successfully corrected 5 transient error frames, boosting the overall accuracy from a raw value of 97.71% to 99.14% and raising the MAA from its raw value of 97.25% to 98.28%. Notably, during the 104 s effective recording duration, the theoretical expected data volume was 520 frames based on a 200 ms sliding window step size, yielding a rate of 5 frames per second. However, due to network packet loss during UDP transmission and thread blocking caused by online computational latency, 350 prediction frames were actually received and processed, accounting for approximately 67.3% of the theoretically expected frame count. Nevertheless, these results clearly demonstrate that the post-processing strategy effectively stabilis es online predictions under real-time constraints. The interface was also applied to control a robotic hand [Figure 5D]. The system is equipped with an ESP32-S3 as the computing centre, which has only 512 KiB of SRAM [Figure 5E]. The full 34-gesture LR model contains 2,210 double-precision parameters occupying 17.68 kB of memory [Table S2]. Consequently, both training and inference are extremely fast. Due to the mechanical hand's degrees of freedom, we tested only 20 gestures, including single-finger, multi-finger, and one resting gesture [Movie S2]. Accordingly, the model configured for this 20-gesture robotic hand task requires only 1,300 double-precision parameters, taking up just 10.40 kB of memory [Table S2]. Each gesture was independently tested for five trials during the robotic hand control experiment. Some prediction errors occurred between similar gestures, such as the "On" and "IF" gestures, whose main motion patterns both involve extension of the index finger. Furthermore, post-processed sliding-window results from five repeated trials for each gesture were aggregated to construct the overall confusion matrix [Figure 5F], yielding an overall recognition accuracy of 98.99%. However, because sample sizes across gestures could not be strictly balanced during continuous online execution, and the rest state naturally predominated in sample count, the MAA of 98.21% serves as a more reliable and unskewed metric for evaluating multi-class performance.

Regarding online computation performance of 34-gesture recognition, the LR model requires 2,210 parameters and 4.39 kFLOPs. Measured on an ESP32-S3 microcontroller operating at 160 MHz, on-device execution achieves a single-sample inference latency of 0.28 ms and a maximum processing throughput of 3,563.8 Hz. Offline evaluations across 15 subjects demonstrate that window sizes below 150 ms fail to yield sufficient decoding accuracy, whereas windows exceeding 250 ms introduce excessive latency when combined with post-processing [Figure S7]. Consequently, a 200 ms data buffering window was selected as an optimal compromise for online signal acquisition. Furthermore, a consecutive consistency post-processing strategy serves as a confidence window, introducing an intentional decision delay of 200 ms to 400 ms to enhance operational stability.

High recognition accuracy on the resource-constrained ESP32-S3 platform fully underscores the practical viability of our HMI-tailored neuromotor interface.

When paired with high-quality physical data, our LR model with merely 2,210 parameters outperforms representative lightweight spatio-temporal deep networks [Table S3]. This remarkable result demonstrates that optimising physical data quality directly enables lightweight, high-precision decoding, fundamentally eliminating the reliance on computationally intensive deep learning models. Such efficiency renders our method ideal for edge computing applications, including smart wearables, continuous health monitoring, and advanced prosthetic control.

## Discussion

In this study, we demonstrate that adopting a data-centric paradigm effectively solves the algorithm deployment bottleneck on resource-constrained edge devices in neuromotor interfaces [27]. By upgrading physiological signal acquisition quality, an ultra-compact model (2,210 parameters, 17.68 kB) saw its dexterous multi-intent decoding accuracy surge from ~45% under traditional data configurations to 94.36%, enabling successful real-time online validation on edge hardware. Visualizations via t-SNE and DBI analysis [Figure 4] reveal that joint spatio-temporal high-fidelity acquisition (enhanced TSR, elevated SSR, and optimized CL) captures stable signal fluctuations and localized muscle activation patterns. This fundamentally enlarges inter-class separability while tightening intra-class scatter. Consequently, within our data-centric paradigm, the algorithmic requirement shifts from resolving complex, entangled boundaries to efficiently leveraging the intrinsic separability inherent in high-fidelity physiological signals.

Furthermore, to systematically evaluate our data-centric paradigm, we ground our empirical statistics, including ANOVA and clustering metrics, in the physical underpinnings of neuromotor control. Statistical analysis reveals that TSR and SSR exert a dominant main effect on decoding accuracy [Figure 3F]. From a physiological standpoint, elevated TSR preserves the phase coherence and morphological integrity of individual Motor Unit Action Potentials (MUAPs) during high-frequency neural firing, effectively eliminating temporal aliasing. Concurrently, an enhanced SSR resolves subtle spatial propagation gradients and localised co-activation patterns across adjacent muscle fascicles. Such physiological high-fidelity directly induces quantifiable changes within the feature geometry space. Results from our t-SNE visualisations, DBI, and FSI analyses demonstrate that joint spatio-temporal high-fidelity acquisition expands inter-class distances while tightening intra-class scatter, thereby establishing substantially wider decision margins [Figure 4]. Under high-TSR and high-SSR conditions, the performance gap between lightweight linear models and complex non-linear algorithms narrowed dramatically, with linear models even outperforming their non-linear counterparts under specific high-fidelity setups [Figure S5]. Mechanistically, when hardware-level signal acquisition yields intrinsically linearly separable features, lightweight algorithms no longer require high-capacity non-linear architectures to untangle complex class boundaries. Instead, compact linear models natively establish smooth, robust decision surfaces, effectively leveraging the inherent separability of the high-fidelity signals. This statistical and geometric convergence directly substantiates the core thesis of our data-centric paradigm: capturing high-fidelity signals at the physical interface relieves the computational burden on downstream algorithms, making compact models fully capable of dexterous multi-intent recognition.

This study also offers valuable insights for research across multiple domains. First, our work provides a systematic reference for parameter configuration in the design of sEMG-based flexible electronic skin[38,39], supporting the rational development of high-performance wearable neuromotor interfaces [40]. Second, the data-centric paradigm offers practical strategies for resource-constrained applications, including dexterous prosthetic and exoskeleton control[41,42], human performance augmentation[43], wearable health monitoring[44], and other scenarios where traditional model-centric approaches are limited. Third, the data-centric paradigm may inspire new approaches for other neuromotor interfaces that do not have strict edge-deployment requirements, such as brain–machine interfaces [45–47]. By providing a predictable pathway for algorithm optimisation toward lightweight designs, it can enhance the feasibility of long-term, in vivo, and flexible neuromotor applications. Lastly, it provides a low-complexity, streamlined framework and hardware solutions for building dexterous neuromuscular interfaces, breaking through the technical barriers for researchers beyond conventional HMI and neuroengineering fields. Its simplicity and accessibility enable rapid implementation of precise neuromotor control, supporting a wide range of applications, i.e., immersive AR/VR interaction, teleoperated robotics [48], human–robot collaboration [49], and human–vehicle interfaces [50]. By combining methodological guidance with practical implementation, this work fosters interdisciplinary exploration at the intersection of biomechanics, artificial intelligence, and human–machine interaction. Overall, it highlights the broad relevance and applicability of the data-centric approach.

### Limitations

While this study yields valuable insights, several limitations highlight important directions for further research within our data-centric framework. Currently, our findings are based on a small cohort of healthy, right-handed young adults recorded in controlled, single-day controlled laboratory environment. To systematically validate this paradigm, our immediate priority is to scale up within-subject evaluations across a broader and more diverse population, including left-handed individuals, older adults, and individuals with motor impairments, before confronting the inherently tougher challenge of cross-subject decoding. Beyond user diversity, wearable HMI systems in the field must constantly contend with complex real-world perturbations. Whether a data-centric approach can truly buffer against these classic bottlenecks remains an open question. We hypothesise that high-fidelity signal acquisition constructs inherently stable, well-separated feature manifolds with wider decision margins, potentially making the system less vulnerable to physical perturbations such as electrode displacement, day-to-day signal drift, sweat accumulation, long-term re-wearing, and motion artefacts. Testing this hypothesis forms a core part of our ongoing work. Additionally, while the Feature Separability Index provides a practical tool for evaluating intrinsic signal quality, its validation in this work remains restricted to a select set of conventional classifiers. Future studies will explore whether FSI can consistently predict decoding ceilings across broader algorithmic families, including deeper neural architectures. Finally, we aim to push this data-centric philosophy beyond static gesture recognition into continuous, dynamic movement decoding, while incorporating closed-loop feedback mechanisms [51] for more intuitive and responsive control.

## Conclusion

In this study, we explored a data-centric paradigm to address the practical challenges of deploying complex multi-intent recognition algorithms in neuromotor interfaces under limited computational resources. By introducing this framework and utilizing a novel wireless bioelectrical signal acquisition system, our empirical evaluations demonstrate that improving hardware-level signal quality can significantly unlock the performance potential of highly compact models. For a dexterous multi-intent decoding task, providing high-quality sEMG data—achieved through elevated TSR, enhanced SSR, and optimized CL—elevated the offline decoding accuracy of an ultra-compact model from ~45% to 94.36%. With only 2,210 parameters (17.68 kB memory footprint, with potential for further compression via quantization or reduced precision), this model showed initial feasibility for real-time online validation on resource-constrained edge devices. Dimensionality reduction and cluster analyses suggest that spatio-temporal high-fidelity signal acquisition effectively captures steadier signal fluctuations, subtle spatial differences, and large-scale muscle activation patterns. Within our experimental scope, this high fidelity helps construct more distinguishable feature distributions for diverse intentions while maintaining uniformity for identical ones. Consequently, these findings indicate that under a data-centric design paradigm, the algorithmic requirement can shift toward efficiently leveraging the intrinsic separability provided by high-quality physiological signals, offering a practical reference for designing lightweight and resource-efficient neuromotor interfaces.

## Acknowledgments

The authors would like to thank all volunteers for their participation in this study.

## Authors' contributions

Made substantial contributions to conception and design of the study and performed data analysis and interpretation: Jiaxuan Li, Liming Shu, Di Wu;
Performed material preparation and device development: Jiaxuan Li, Di Wu, Liming Shu;
Performed data acquisition, as well as provided administrative, technical, and material support: Jiaxuan Li, Jianhua Liu, Jinnuo Li, Yuxin Zhao, Xiao Zhang, Xiang Li;
Undertook supervision, funding acquisition and project administration: Liming Shu, Di Wu, Zhenzhi Ying, Changsheng Dai

## Availability of data and materials

The data that support the findings of this study are available from the corresponding author upon reasonable request. All circuit hardware designs, including circuit schematics, PCB layouts, and firmware logic, were independently developed by our team. Standard commercial off-the-shelf electronic components were utilized for circuit implementation, and detailed hardware specifications are available upon reasonable request.

## AI and AI-assisted Tools Statement

During the preparation of this work, the authors used Gemini (Gemini 2.0) for language polishing, grammar refinement, and assisting in generating specific graphical elements, including: Graphical Abstract: The arm schematic, electronic skin, constrained edge device, prosthetics, data glove, and AR/VR glasses icons; Figure 1: The two arm schematics and four gesture elements in Figure 1A; the human silhouette, brain, neuron cells, and robotic hand schematics in Figure 1B; Figure 2: The reference electrode schematic in Figure 2A; Figure 3 & Figure 5: The arm schematic in Figure 3C, and the subject schematic in Figure 5B; Supplementary Materials: The arm muscle distribution schematic in Figure S2.

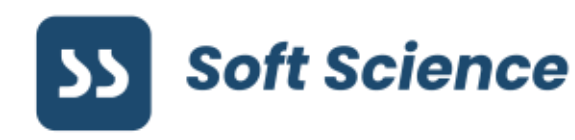

The authors reviewed and edited all AI-generated content and accept full responsibility for the published material.

## Financial support and sponsorship

This work was supported by the National Natural Science Foundation of China (No. 52475242) and the Fundamental Research Funds for the Central Universities (No. DUT25YG249). The funders had no involvement in study design, data processing and manuscript composition.

## Conflicts of interest

All authors declared that there are no conflicts of interest.

## Ethical approval and consent to participate

## Declarations

### Ethical Approval and Consent to Participate

This study was approved by the Ethics Committee of Dalian University of Technology (approval number: DUTSME230406-01) and conducted in accordance with the Declaration of Helsinki. Written informed consent to participate was obtained from all individual participants prior to experimentation.

### Consent for Publication

Written informed consent for publication has been acquired from all participants. Specifically, all participants whose identifiable facial images are presented in this manuscript were fully informed that these images would be published and provided written consent for their publication.

## Consent for publication

Not applicable